\pdfoutput=1
\documentclass[conference, 10pt]{IEEEtran}
\IEEEoverridecommandlockouts
\usepackage{placeins}
\usepackage{cite}
\usepackage{amsmath,amssymb,amsfonts}
\usepackage{algorithmic}
\usepackage{graphicx}
\usepackage{textcomp}
\usepackage{xcolor}
\usepackage{url}
\usepackage{booktabs}
\usepackage{bm}
\usepackage{pifont}
\usepackage[colorlinks=true, linkcolor=black, citecolor=black, urlcolor=blue!60!black]{hyperref}

\def\BibTeX{{\rm B\kern-.05em{\sc i\kern-.025em b}\kern-.08em
    T\kern-.1667em\lower.7ex\hbox{E}\kern-.125emX}}

\newif\ifarxiv
\arxivtrue

\begin{document}

\title{Smart Compaction: Predicting Compaction Utility from Lakehouse Table Metadata\textsuperscript{\ding{73}}%
\thanks{\ding{73}\,The views expressed in this paper are solely those of the authors and do not necessarily reflect the views of the European Central Bank, the European System of Central Banks or IBM. All code and data for full replication are available at \url{https://github.com/JannicCutura/smart-compaction}.}
}

\author{\IEEEauthorblockN{Jannic Alexander Cutura}
\IEEEauthorblockA{\textsc{ECB}\textsuperscript{\dag} \&\, \textsc{DSTI}\textsuperscript{\ddag}\\
{\small\color{gray!80!black}\href{mailto:jannic.cutura@dsti.institute}{\texttt{jannic.cutura@dsti.institute}}}}
\and
\IEEEauthorblockN{Subash Prakash}
\IEEEauthorblockA{\textsc{IBM}\textsuperscript{\S}\\
{\small\color{gray!80!black}\href{mailto:subash.prakash@ibm.com}{\texttt{subash.prakash@ibm.com}}}}
}

\maketitle

\renewcommand{\thefootnote}{\fnsymbol{footnote}}
\footnotetext[2]{European Central Bank, Sonnemannstra\ss e 20, 60314 Frankfurt am Main, Germany.}
\footnotetext[3]{DSTI School of Engineering, 59 Rue de l'Abb\'e Groult, 75015 Paris, France.}
\footnotetext[4]{IBM, IBM Campus 1, 71139 Ehningen, Germany.}
\renewcommand{\thefootnote}{\arabic{footnote}}
\ifarxiv
{\let\thefootnote\relax\footnotetext{\copyright~2026 IEEE. Personal use of
this material is permitted. Permission from IEEE must be obtained for all
other uses, in any current or future media, including reprinting/republishing
this material for advertising or promotional purposes, creating new collective
works, for resale or redistribution to servers or lists, or reuse of any
copyrighted component of this work in other works.}}
\fi

\begin{abstract}
Open lakehouse table formats accumulate small data files over time,
which degrades query performance. Deciding when compaction is worthwhile
remains threshold-driven, but which metadata features actually determine
compaction utility is not well understood. We present an open simulation
framework that generates 2{,}376 Apache Iceberg tables spanning three
orders of magnitude in file size, extracts 17 metadata features from
manifest files without reading data, and trains XGBoost to predict the
continuous file-reduction ratio ($R^2 = 0.998$, RMSE~$= 0.013$). The
binary compaction decision turns out to be trivially separable by a
single partition-level threshold
(\texttt{max\_files\_per\_partition}~$> 4$), requiring no learned model.
Cross-schema validation on 96 TPC-H tables confirms generalisation
without retraining ($R^2 = 0.976$). A query benchmark reveals that
compaction benefits metadata-heavy queries but can slow full-scan
aggregations by reducing task parallelism.
All code and data are publicly available.
\end{abstract}


\section{Introduction}

The lakehouse architecture~\cite{armbrust2021lakehouse} unifies data warehousing
and data lake workloads through open table formats like Apache
Iceberg~\cite{iceberg_spec}, Delta Lake~\cite{armbrust2020delta}, and Apache
Hudi~\cite{hudi_compaction}. Each transaction appends immutable columnar
files, buying snapshot isolation at the cost of progressive file
accumulation.

This yields the \emph{small file problem}: file counts grow without bound,
inflating query planning latency, per-request object-store costs, and scan
overhead. At scale, catalogs can accumulate millions of undersized
files~\cite{gruenheid2025autocomp}. \emph{Compaction} rewrites them into
fewer target-sized files via bin-packing, sort, or z-order strategies, but
is itself costly in compute and I/O. The question is therefore not
\emph{whether} to compact but \emph{when} the file reduction justifies the
cost.

Currently, all open-source compaction scheduling is threshold-driven:
Iceberg's \texttt{rewrite\_data\_files} is purely imperative~\cite{iceberg_spec},
Delta Lake triggers on small-file counts~\cite{armbrust2020delta}, and Hudi
uses rule-based triggers~\cite{hudi_compaction}.
AutoComp~\cite{gruenheid2025autocomp}, the most comprehensive automated system
to date, uses heuristics and multi-objective Pareto ranking over 35{,}000
Iceberg tables at LinkedIn; its authors explicitly identify ML-based prediction
as future work. Proprietary systems (Databricks Predictive
Optimization~\cite{databricks_predictive_optimization}, Snowflake automatic
clustering~\cite{snowflake_autoclustering}, AWS S3
Tables~\cite{aws_s3tables_compaction}) are not open or reproducible.
No published study has examined \emph{which} table metadata features actually
determine compaction utility.

This paper addresses the gap. We generate 2{,}376 Iceberg tables spanning three
orders of magnitude in file size, extract 17 metadata features from manifest
files without reading data, and evaluate how well these features predict
bin-packing compaction outcomes. We make three contributions:
\begin{enumerate}
\item A reproducible simulation framework and open dataset of 2{,}376 Iceberg
      tables with heterogeneous write patterns, Zipfian skew, and
      file-size heterogeneity---the last being essential to avoid a
      \emph{tautology trap} in which uniformly small files make outcomes
      deterministic.
\item An XGBoost regressor that predicts the continuous file-reduction
      ratio ($R^2 = 0.998$, robust across prevalence levels, 30-seed
      bootstrap splits, and feature subsets). The binary compaction
      decision, by contrast, reduces to a deterministic threshold
      (\texttt{max\_files\_per\_partition}~$>$~4) and does not require
      a learned model.
\item Cross-schema validation on 96 TPC-H tables confirming
      generalisation without retraining ($R^2 = 0.976$), plus a query
      benchmark revealing that compaction benefits metadata-only
      queries but can slow full-scan aggregations by reducing task
      parallelism.
\end{enumerate}

\subsection{Compaction in LSM-Trees}

LSM-tree research has established merge-policy cost
models~\cite{dayan2017monkey, dayan2018dostoevsky}, quantified the
$63\times$ data movement overhead~\cite{sarkar2021compaction}, and
applied RL~\cite{mo2023ruskey} and dynamic
programming~\cite{ecotune2025} to timing decisions.
Lakehouse tables differ fundamentally: flat columnar files on object
stores with no level hierarchy, where file proliferation rather than
write amplification drives cost. We share the core question ---
\emph{when is reorganising stored data worth the compute cost?} ---
but answer it with metadata-driven prediction rather than policy
optimisation.

\subsection{Automated Database Tuning and ML for Systems}

Database auto-tuning has progressed from rule-based
advisors~\cite{chaudhuri2007selftuning} through
OtterTune~\cite{vanaken2017ottertune} (Gaussian processes) and
CDBTune~\cite{zhang2019cdbtune} (deep RL) to
NoisePage's~\cite{pavlo2021noisepage} forecast--model--plan framework,
which our work instantiates: metadata features forecast compaction
utility, and the binary decision is the action plan.
In the broader ``ML for systems'' movement~\cite{zhou2022dbmeetsai},
Bao~\cite{marcus2021bao} steers query optimisation through learned hint
selection and Qd-tree~\cite{yang2020qdtree} learns workload-aware
layouts. Yet no published work applies ML to predict the utility of a
storage maintenance operation in the lakehouse.
AutoComp~\cite{gruenheid2025autocomp} is the closest predecessor,
relying on heuristic trait computation rather than learned models.

\section{Methodology}

\subsection{Problem Formulation}

We define the \emph{compaction utility} of a table as the file count
reduction ratio achieved by running Iceberg's \texttt{rewrite\_data\_files}
with bin-packing:
\begin{equation}
  r_{\mathrm{fc}} = 1 - \frac{n_{\mathrm{after}}}{n_{\mathrm{before}}}
  \label{eq:rfc}
\end{equation}
where $n_{\mathrm{before}}$ and $n_{\mathrm{after}}$ are the number of data
files before and after compaction, respectively. A value of
$r_{\mathrm{fc}} \approx 1$ indicates high compaction benefit;
$r_{\mathrm{fc}} \approx 0$ means that compaction produces negligible file
reduction and the associated I/O cost is wasted.

We address two complementary tasks:
\begin{enumerate}
\item \textbf{Classification.} Predict whether a table \emph{needs compaction},
      defined as whether Iceberg's bin-packing actually rewrites any data
      files ($n_{\mathrm{rewritten}} > 0$). This label captures compaction
      \emph{activity} rather than \emph{utility}; however, the distribution
      is bimodal: 99.5\% of positive labels have $r_{\mathrm{fc}} > 0.5$
      (only 11 of 2{,}376 tables lie in $(0, 0.5]$), so activity coincides
      with meaningful file reduction in practice.
\item \textbf{Regression.} Predict the continuous reduction ratio
      $r_{\mathrm{fc}} \in [0, 1]$ to enable cost-aware scheduling.
      This is the paper's primary prediction task.
\end{enumerate}

\subsection{Simulation Framework}

Production tables have unknown ground-truth compaction utility, so we construct a
simulation that generates Iceberg tables with controlled characteristics,
compacts them, and records the outcome.

\subsubsection{Parameter Grid}
Table~\ref{tab:params} lists seven configuration axes spanning 2{,}376 combinations.
The axes cover three orders of magnitude in file size (8\,KB to 128\,MB),%
\footnote{The \texttt{file\_size\_target\_kb} column in Table~\ref{tab:params}
specifies the Spark write-side parameter. Actual on-disk file sizes differ
because the heterogeneous batch pattern (Section~\ref{sec:hetero}) uses an $8\times$
smaller target for even-indexed batches, creating within-table variance.}
partition cardinalities from 1 (unpartitioned) to 100, write parallelism of 1 or
5 writers, append frequencies from 1 to 20 batches, and uniform vs.\ Zipfian
($s{=}1$) partition skew. At 50 UUID columns (${\sim}$1.5\,KB/row in Parquet),
files can reach the 128\,MB compaction target at scale.

\subsubsection{Heterogeneous Write Patterns}
\label{sec:hetero}
To model production heterogeneity, the first batch receives 50\% of rows
(initial load) while remaining batches split the rest equally. Even-indexed
batches use a file-size target $8\times$ smaller than the configured value,
creating within-table variance that breaks the tautology.

\subsubsection{Compaction Procedure}
We invoke Iceberg's \texttt{rewrite\_data\_files} with bin-packing: target
128\,MB, minimum 96\,MB (75\%), maximum 230\,MB (180\%). Files in
$[96, 230]$\,MB are not rewritten, matching default production configuration.
\begin{table}[t]
\setlength{\abovecaptionskip}{4pt}
\setlength{\belowcaptionskip}{-4pt}
\centering
\caption{Simulation Parameter Grid}
\label{tab:params}
\footnotesize
\begin{tabular}{lp{1.8cm}p{2.8cm}}
\toprule
\textbf{Parameter} & \textbf{Description} & \textbf{Values} \\
\midrule
\texttt{num\_rows} & Total rows per table & 10K, 100K, 1M \\
\texttt{num\_columns} & Payload columns (excl.\ part.\ key) & 5, 50 \\
\texttt{num\_partitions} & Number of partitions (1\,=\,none) & 1, 10, 100 \\
\texttt{num\_writers} & Writers per batch & 1, 5 \\
\texttt{num\_write\_batches} & Append rounds & 1, 5, 20 \\
\texttt{file\_size\_target\_kb} & Target file size (KB) & 8, 128, 512, 2K, 4K, 8K, 16K, 24K, 32K, 65K, 131K \\
\texttt{partition\_skew} & Skew profile & uniform, zipf \\
\midrule
\multicolumn{3}{l}{\textit{Full grid: 2376 configurations}} \\
\bottomrule
\end{tabular}
\end{table}

\FloatBarrier

\begin{figure}[!ht]
\centerline{\includegraphics[width=\columnwidth]{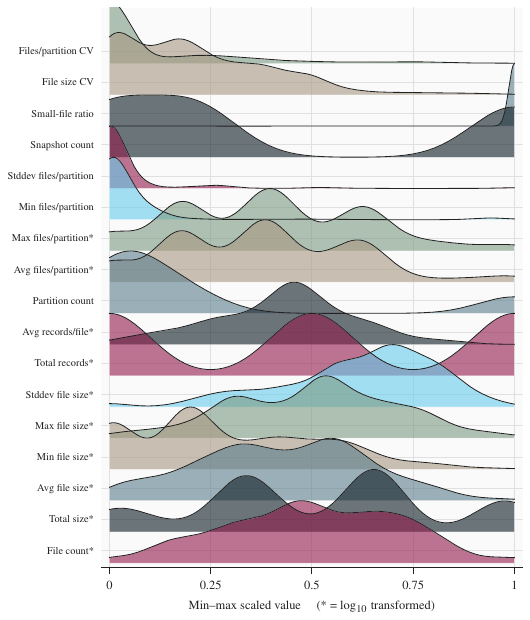}}
\caption{Ridge plot of the 17 input features, each min--max scaled to
  $[0,1]$ for visual comparability.  Features marked with an asterisk are
  $\log_{10}$-transformed before scaling.  Each row shows the kernel density
  estimate of one feature across all 2{,}376 tables.}
\label{fig:violins}
\end{figure}

\subsection{Feature Engineering}

We extract 17 observable features exclusively from Iceberg metadata (manifest
files, snapshot history, and partition statistics) without reading any data
files.  This makes feature extraction fast and non-intrusive in production.
The feature vector comprises eight file-level statistics (\texttt{file\_count},
\texttt{total\_size\_bytes}, average/minimum/maximum/standard-deviation of file
size, \texttt{total\_records}, \texttt{avg\_records\_per\_file}), five
partition-level statistics (\texttt{num\_partitions\_actual},
average/maximum/minimum/standard-deviation of files per partition), one
snapshot count (\texttt{num\_snapshots}), and three derived ratios:
\texttt{small\_file\_ratio} (fraction of files below the 96~MB compaction
threshold), \texttt{file\_size\_cv} (coefficient of variation of file sizes),
and \texttt{files\_per\_partition\_cv} (coefficient of variation of files per
partition).  These ratios capture the fragmentation and heterogeneity that
drive compaction decisions.
Note that \texttt{small\_file\_ratio} uses the same 96\,MB threshold that
Iceberg's bin-packing filter uses to select files for rewriting. This
creates a potential target-leakage concern; we discuss its impact in
Section~\ref{sec:discussion} and show via ablation
(Section~\ref{sec:ablation}) that it is not a top predictor and
removing it does not degrade model performance.

Fig.~\ref{fig:violins} shows the distribution of each feature, split by
compaction outcome.

\subsection{Model Selection}

We train four models on the 17-feature vector: an XGBoost
classifier~\cite{chen2016xgboost} and logistic-regression baseline for the
binary \texttt{needs\_compaction} task, and an XGBoost regressor and OLS
baseline for continuous $r_\mathrm{fc}$. All use an 80/20 stratified
train/test split with 5-fold cross-validation. Classification is evaluated
via accuracy, F1, and ROC-AUC; regression via RMSE and $R^2$.

\section{Experimental Evaluation}

\subsection{Dataset Overview}

The simulation produces 2{,}376 tables. Of these, 2{,}121 (89.3\%) have at
least one rewritten data file and are labelled \texttt{needs\_compaction}~$= 1$.
The file-reduction ratio ranges from 0.000 to 0.999 (mean~$= 0.818$,
median~$= 0.950$, $\sigma = 0.294$), exhibiting a heavy left tail of
tables that gain little from compaction.

\subsection{Binary Compaction Decision}
\label{sec:binary}

Before addressing the regression task, we establish that the binary
compaction decision does not require a learned model.
XGBoost achieves perfect classification (F1~$= 1.000$,
AUC~$= 1.000$) on the held-out test set ($n = 476$), but so does
the trivial threshold
\texttt{max\_files\_per\_partition}~$> 4$.
A threshold sweep reveals the mechanics: under bin-packing
(target 128\,MB, min 96\,MB), every partition with~$\geq 5$ files
contains at least one eligible for rewriting, so the boundary
falls exactly at $k = 4$.
This is a deterministic property of the compaction filter, not a
learned pattern, and the threshold transfers perfectly to TPC-H
(Section~\ref{sec:tpch}).

\begin{figure}[b]
\centerline{\includegraphics[width=\columnwidth]{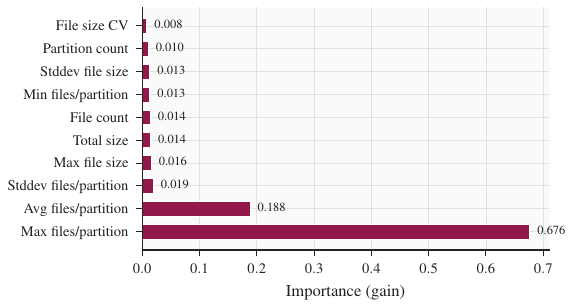}}
\caption{XGBoost feature importance (gain). Two partition-level features
  dominate: 86.4\% of classifier gain, 95.2\% of regressor gain.}
\label{fig:importance}
\end{figure}

Feature importance analysis (Fig.~\ref{fig:importance}) confirms
extreme concentration: \texttt{max\_files\_per\_partition}
(gain~$= 0.676$) and \texttt{avg\_files\_per\_partition}
(gain~$= 0.188$) account for 86.4\% of classifier gain
and 95.2\% of regressor gain. Partition-level file counts
dominate both tasks.

\subsection{Regression}

\begin{figure}[b]
\centerline{\includegraphics[width=\columnwidth]{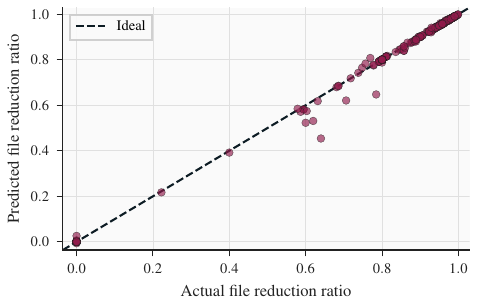}}
\caption{Predicted vs.\ actual file reduction ratio on the held-out test
  set. Points near the dashed line indicate accurate predictions.}
\label{fig:scatter}
\end{figure}

The continuous prediction task is where ML adds genuine value.
Table~\ref{tab:reg} summarises regressor
performance.  XGBoost achieves $R^2 = 0.998$ (RMSE~$= 0.013$,
MAE~$= 0.002$), with a bootstrap 95\% CI of $R^2 \in [0.996, 1.000]$
(Section~\ref{sec:confidence}), while OLS captures only modest variance ($R^2 = 0.361$),
confirming that a linear model is insufficient for this task.
Fig.~\ref{fig:scatter} plots predicted vs.\ actual reduction ratios.

The OLS shortfall is explained by the nonlinear relationship between
files-per-partition and the reduction ratio: the bin-packing target
file size creates a saturation curve: once partitions contain enough
small files to fill a target file, additional files yield diminishing
marginal reduction. XGBoost's tree splits capture this inflection
naturally; a linear model cannot.

The Spearman correlation between
\texttt{avg\_files\_per\_partition} and $r_\mathrm{fc}$ is
$\rho = 0.965$, confirming that the relationship
is monotonic but not linear.

\begin{table}[tb]
\setlength{\abovecaptionskip}{4pt}
\setlength{\belowcaptionskip}{-6pt}
\centering
\caption{Regression results on the held-out test set.}
\label{tab:reg}
\begin{tabular}{lccc}
\toprule
Model & RMSE & MAE & $R^2$ \\
\midrule
XGBoost        & 0.013 & 0.002 & 0.998 \\
Linear Reg.    & 0.236 & 0.159 & 0.361 \\
\bottomrule
\end{tabular}
\end{table}

\textbf{Per-quantile error analysis.}
Because the $r_{\mathrm{fc}}$ distribution is
heavily right-skewed (median~$= 0.950$, mean~$= 0.818$), the global
RMSE of~0.013 could mask larger errors in sparsely populated bins.
Table~\ref{tab:quantile} stratifies the test set by $r_{\mathrm{fc}}$
range. The hardest bin is $(0.5,\,0.8]$ (RMSE~$= 0.027$,
max$|e| = 0.186$), where the bin-packing saturation curve is steepest
and small feature changes cause large shifts in reduction ratio.
Even in this worst case, MAE remains below~0.008, and the remaining
bins have RMSE~$\leq 0.007$. The model's accuracy is therefore
consistent across the full target range.

\begin{table}[tb]
\setlength{\abovecaptionskip}{4pt}
\setlength{\belowcaptionskip}{-6pt}
\centering
\caption{Per-bin regression error on the held-out test set.}
\label{tab:quantile}
\begin{tabular}{lcrrr}
\toprule
$r_{\mathrm{fc}}$ bin & $n$ & RMSE & MAE & Max$|e|$ \\
\midrule
$r_{fc}=0$ & 51 & 0.0039 & 0.0011 & 0.026 \\
$(0,\,0.5]$ & 2 & 0.0066 & 0.0064 & 0.008 \\
$(0.5,\,0.8]$ & 106 & 0.0272 & 0.0074 & 0.186 \\
$(0.8,\,0.95]$ & 201 & 0.0027 & 0.0011 & 0.018 \\
$(0.95,\,1]$ & 235 & 0.0010 & 0.0005 & 0.008 \\
\bottomrule
\end{tabular}

\end{table}

\subsection{Cross-Schema Generalisation}
\label{sec:tpch}

The preceding analysis trains and tests on tables sharing the same synthetic
schema (50 UUID columns). To test whether the learned decision boundary
transfers across data types and schemas, we evaluate the trained models without
retraining on 96 Iceberg tables derived from six TPC-H~\cite{tpch2024} schemas (lineitem,
orders, customer, part, partsupp, supplier), the same benchmark used by
Jain et al.~\cite{jain2023analyzing} for cross-format lakehouse evaluation
and by LST-Bench~\cite{camachorodriguez2024lstbench} for log-structured table
benchmarking. These tables use realistic data types (strings, dates,
decimals, and integers) with 7--16 columns per schema instead of the
uniform 50-column UUID payload, and are generated at approximately TPC-H
scale factor~10 (468.8\,M total rows, 41{,}282 files, 15\,GB
pre-compaction).

To ensure non-trivial compaction outcomes, row counts are scaled so that files
span Iceberg's default 96\,MB compaction threshold~\cite{iceberg_spec}. The
resulting label balance is 79 tables needing compaction (82.3\%) vs.\ 17 that
do not.

Table~\ref{tab:tpch} summarises the results.
The regressor achieves $R^2 = 0.976$ (RMSE~$= 0.058$), a moderate
degradation from the training distribution ($R^2 = 0.998$) that
confirms the model captures transferable structure rather than
artefacts of the synthetic schema.
The $k{=}4$ threshold transfers to TPC-H at 97.9\% accuracy,
with only two false negatives---both borderline cases (3~files
at~$\sim$50\,MB average, $r_\mathrm{fc} = 0.333$) where the
compactor merged 3~files into~2.

\subsection{Feature Ablation}
\label{sec:ablation}

To determine how many features are necessary and where information
concentrates, we conduct two complementary ablation studies.

\begin{figure}[tb]
\centerline{\includegraphics[width=\columnwidth]{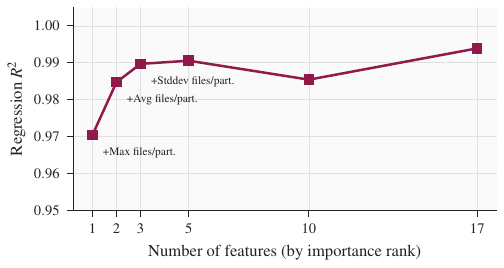}}
\caption{Regression $R^2$ vs.\ the number of top-$k$ features.
  Two features capture 98.5\% of regression variance.}
\label{fig:ablation}
\end{figure}

\textbf{Progressive top-$k$.} We train XGBoost using only the top-$k$
features ranked by gain importance from the full model
(Fig.~\ref{fig:ablation}). A single feature
(\texttt{max\_files\_per\_partition}) achieves
$R^2 = 0.970$. Adding \texttt{avg\_files\_per\_partition} raises
$R^2$ to~0.985; the top-3 features reach~0.990. Beyond five features
returns diminish, and ten features slightly degrade regression
($R^2 = 0.985$), consistent with mild overfitting from irrelevant
splits.

\textbf{Leave-one-out.} All single-feature removals maintain
$R^2 \geq 0.988$, showing no feature is individually indispensable
when the remaining 16 are available.

To formally quantify the marginal value of features~3--17, we bootstrap
50 random train/test splits and compute the paired difference
$\Delta R^2 = R^2_{k=17} - R^2_{k=2}$ on each.
The mean $\Delta R^2 = 0.023$ with 95\% CI $[0.004, 0.044]$;
the CI excludes zero, confirming a statistically significant
but practically small gain.

These results empirically confirm the feature dominance observed in
Section~\ref{sec:binary}: partition-level file counts carry nearly all the
predictive signal. The 15 auxiliary features contribute a
statistically significant but small marginal improvement
($\Delta R^2 \approx 0.02$), serving primarily as partial proxies
when file count features are noisy or unavailable.

\subsection{Hyperparameter Sensitivity and Model Families}
\label{sec:sensitivity}

We evaluate robustness to XGBoost configuration by sweeping
\texttt{max\_depth}~$\in \{2, 4, 6, 8\}$ and
\texttt{n\_estimators}~$\in \{10, 50, 200, 500\}$
(16~configurations). Regression is sensitive to capacity:
depth-2 with 10~trees
yields $R^2 = 0.852$, but depth-2 with 50~trees already reaches
$R^2 = 0.994$. Configurations with depth~$\geq 4$ and $\geq 50$
estimators saturate at $R^2 \geq 0.997$.

To verify that the finding is not XGBoost-specific, we compare four
model families (Table~\ref{tab:modelfamily}). Decision trees and
random forests achieve near-perfect metrics, confirming that any
tree-based model suffices. The linear baselines perform markedly
worse on both tasks---logistic regression reaches only
F1~$= 0.885$ on the binary decision, and OLS attains
$R^2 = 0.361$ on the regression target---underscoring the
nonlinear structure of $r_\mathrm{fc}$.

\begin{table}[tb]
\centering
\caption{Model family comparison (default hyperparameters).}
\label{tab:modelfamily}
\begin{tabular}{lcccc}
\toprule
Model & Clf Acc. & Clf F1 & Reg $R^2$ & Reg RMSE \\
\midrule
XGBoost (ours)    & 1.000 & 1.000 & 0.998 & 0.013 \\
Random Forest     & 1.000 & 1.000 & 0.998 & 0.013 \\
Decision Tree     & 0.996 & 0.998 & 0.997 & 0.016 \\
Linear\textsuperscript{$\dagger$} & 0.815 & 0.885 & 0.361 & 0.236 \\
\bottomrule
\end{tabular}
\\[2pt]
{\raggedright\footnotesize
\textsuperscript{$\dagger$}Logistic regression for the classification
columns, ordinary least squares for the regression columns.\par}
\end{table}

\subsection{Statistical Confidence and Prevalence Robustness}
\label{sec:confidence}

\textbf{Bootstrap confidence intervals.} To guard against split-dependent
results, we repeat the 80/20 stratified split 30~times with different
random seeds and report 95\% percentile bootstrap CIs.
The regressor achieves $R^2 = 0.999$
$[0.996, 1.000]$ and RMSE~$= 0.009$ $[0.004, 0.019]$.

\textbf{Prevalence sensitivity.} The 89.3\% positive prevalence
raises legitimate concern that metrics are inflated by class
imbalance. We downsample the majority class to create balanced
evaluation sets at 70\%, 50\%, and 30\% positive prevalence
(Table~\ref{tab:prevalence}).
Regression
$R^2$ stays above~0.995 at all levels, confirming that model
performance is not an artefact of class imbalance.
The $k{=}4$ threshold degrades slightly at
lower prevalence (97.5\% at 30\%) but remains competitive for the
binary decision.

\begin{table}[tb]
\centering
\caption{Prevalence robustness evaluation.}
\label{tab:prevalence}
\begin{tabular}{ccccc}
\toprule
Prev. & $n$ & XGB Acc. & $k{=}4$ Acc. & Reg $R^2$ \\
\midrule
89.3\% & 2{,}376 & 1.000 & 1.000 & 0.998 \\
70\%   &    849 & 1.000 & 0.989 & 0.995 \\
50\%   &    510 & 1.000 & 0.982 & 0.998 \\
30\%   &    364 & 1.000 & 0.975 & 0.997 \\
\bottomrule
\end{tabular}
\\[2pt]
{\raggedright\footnotesize
XGBoost and the threshold baseline ($k{=}4$) evaluated on downsampled
datasets; regression uses XGBoost.\par}
\end{table}

\begin{table}[b]
\centering
\caption{Cross-schema validation on 96 TPC-H tables (no retraining).}
\label{tab:tpch}
\smallskip
\footnotesize
\textit{(a) Aggregate results}\\[2pt]
\begin{tabular}{lcccc}
\toprule
\textbf{Task / Model} & \textbf{Acc.} & \textbf{F1} & \textbf{RMSE} & $\bm{R^2}$ \\
\midrule
\multicolumn{5}{l}{\textit{Classification}} \\
\quad XGBoost         & 0.979 & 0.987 & --- & --- \\
\quad Threshold ($k{=}4$) & 0.979 & 0.987 & --- & --- \\
\quad Threshold ($\texttt{sfr}{>}0$) & 0.833 & 0.908 & --- & --- \\
\quad Majority class  & 0.823 & 0.903 & --- & --- \\
\midrule
\multicolumn{5}{l}{\textit{Regression}} \\
\quad XGBoost         & --- & --- & 0.058 & 0.976 \\
\bottomrule
\end{tabular}
\\[6pt]
\textit{(b) Per-schema breakdown}\\[2pt]
\begin{tabular}{lccccc}
\toprule
Schema & $n$ & Clf Acc. & Clf F1 & Reg $R^2$ & Reg RMSE \\
\midrule
lineitem & 16 & 1.000 & 1.000 & 0.997 & 0.019 \\
orders   & 16 & 1.000 & 1.000 & 1.000 & 0.003 \\
part     & 16 & 1.000 & 1.000 & 1.000 & 0.002 \\
supplier & 16 & 1.000 & 1.000 & 1.000 & 0.002 \\
customer & 16 & 0.938 & 0.966 & 0.874 & 0.097 \\
partsupp & 16 & 0.938 & 0.966 & 0.867 & 0.101 \\
\bottomrule
\end{tabular}
\end{table}

Table~\ref{tab:tpch}(b) breaks down performance by TPC-H schema.
Four schemas achieve near-perfect regression
($R^2 > 0.997$). Customer and partsupp show lower metrics
($R^2 \approx 0.87$, $n = 16$ each), likely because these schemas
produce borderline tables near the compaction threshold.

\subsection{Query Performance Impact}
\label{sec:querybench}

To validate that structural compaction utility translates to
downstream benefit, we benchmark query latency before and after
compaction on 52 TPC-H tables using Iceberg time-travel
(\texttt{VERSION AS OF}) to query both snapshots non-destructively.
Each query is run with 1~warmup and 3~timed iterations; we report
the median.

Table~\ref{tab:querybench} summarises the results by query type.
Full-scan \texttt{COUNT(*)} queries benefit consistently from
compaction (median 1.19$\times$ speedup), with higher gains on
tables with more files (lineitem: 1.41$\times$ at $\sim$1{,}200
pre-compaction files). The Spearman correlation between
file-reduction ratio and \texttt{COUNT(*)} speedup is
$\rho = 0.50$ ($p < 0.001$), confirming that metadata overhead
(file listing, Parquet footer reads) dominates this query type.

Filtered aggregation queries show mixed results. The date-filtered
\texttt{orders\_agg} achieves 1.00$\times$ (neutral), while
\texttt{tpch\_q1} on lineitem shows 0.76$\times$ (24\% slower).
Full-scan aggregations without predicates
(\texttt{cust\_nation\_agg}, \texttt{part\_type\_agg}) are
consistently slower after compaction (0.46$\times$ and
0.45$\times$ respectively). We attribute this to reduced task
parallelism: Spark assigns tasks proportional to file count, so
compacting 300 files into~1 reduces parallelism from 300 tasks
to~1, offsetting the metadata benefit for queries that must read
all data. This finding is consistent with observations in
LST-Bench~\cite{camachorodriguez2024lstbench} that compaction
effects are workload-dependent.

\begin{table}[tb]
\setlength{\abovecaptionskip}{4pt}
\setlength{\belowcaptionskip}{-4pt}
\centering
\caption{Query latency speedup after compaction.}
\label{tab:querybench}
\begin{tabular}{lccccc}
\toprule
Query & $n$ & Med.\ & Mean & $\rho$ & $p$ \\
\midrule
\texttt{count\_star}    & 52 & 1.19$\times$ & 1.31$\times$ & 0.50 & $<$0.001 \\
\texttt{orders\_agg}    & 12 & 1.00$\times$ & 1.63$\times$ & 0.89 & $<$0.001 \\
\texttt{tpch\_q1}       & 13 & 0.76$\times$ & 0.82$\times$ & 0.77 & 0.002 \\
\texttt{cust\_nation}   & 15 & 0.46$\times$ & 0.56$\times$ & 0.45 & 0.095 \\
\texttt{part\_type}     & 12 & 0.45$\times$ & 0.47$\times$ & 0.38 & 0.226 \\
\bottomrule
\multicolumn{6}{l}{\scriptsize 52 TPC-H tables; median of 3 timed runs.} \\
\multicolumn{6}{l}{\scriptsize Speedup~$> 1{\times}$ = faster post-compaction.} \\
\end{tabular}
\end{table}

\begin{figure}[b]
\centerline{\includegraphics[width=\columnwidth]{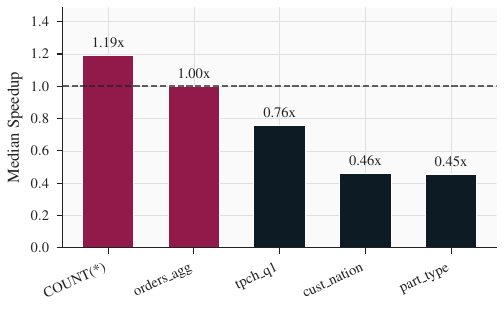}}
\caption{Median query speedup by query type after compaction.}
\label{fig:queryspeedup}
\end{figure}

The overall median speedup across all 104~measurements is
0.97$\times$ (mean 1.08$\times$), indicating that bin-packing
compaction is not universally beneficial for query performance
(Fig.~\ref{fig:queryspeedup}).
This reinforces the value of a learned compaction scheduler: rather
than compacting all tables with small files, a production system
should consider query workload when prioritising compaction, using
the regression prediction as a structural cost signal and the
query-type mix to estimate downstream benefit.

\subsection{Discussion}
\label{sec:discussion}

\subsubsection{Label Definition and Feature Leakage}
The binary label \texttt{needs\_compaction} captures whether the compactor
\emph{rewrote} any files, not whether the resulting reduction was
practically beneficial. This distinction matters in principle, but our
dataset mitigates it: the $r_{\mathrm{fc}}$ distribution is bimodal
(mean~$= 0.818$, median~$= 0.950$), with 99.5\% of positive-label tables
exhibiting $r_{\mathrm{fc}} > 0.5$. A table that triggers compaction
almost always achieves substantial file reduction. Regardless, the
paper's primary contribution is the regression model, which predicts
$r_{\mathrm{fc}}$ directly and does not depend on the binary label.

The feature \texttt{small\_file\_ratio} (fraction of files below 96\,MB)
uses the same threshold as the compaction filter, raising a
target-leakage concern. However, its contribution is negligible in
both tasks: it carries only 0.009 gain in the regressor --- against
0.795 for \texttt{max\_files\_per\_partition} --- and does not enter
the classifier's top ten at all.
The leave-one-out ablation (Section~\ref{sec:ablation}) confirms this:
removing it does not degrade $R^2$ below 0.988. The leakage
is therefore present but inconsequential.

\subsubsection{Operational Cost-Benefit Threshold}
A production scheduler needs more than a predicted $r_{\mathrm{fc}}$---it
needs a threshold below which compaction is not worth the I/O cost.
Our dataset provides empirical guidance: of the 2{,}121 tables that
trigger compaction, only 11 (0.5\%) have $r_{\mathrm{fc}} \leq 0.5$,
and tables with $r_{\mathrm{fc}} \leq 0.1$ eliminate zero files
on average. A natural scheduling rule is therefore: compact only
when the predicted $r_{\mathrm{fc}} > 0.3$, which would avoid the
four lowest-utility compactions while retaining~99.8\% of beneficial
ones. The precise threshold is deployment-specific (it depends on
storage cost, I/O bandwidth, and query patterns), but the regressor
provides the continuous signal needed for cost-aware prioritisation.

\subsubsection{Tautology Trap}
An earlier parameter grid produced only small files (max $\sim$0.4~MB vs.\ the
96~MB threshold), so \emph{every} table compacted completely:
$r_\mathrm{fc} = 1 - 1/N$. XGBoost achieved $R^2 = 0.99$ by recovering this
closed-form identity (Spearman $\rho = 0.996$ between
\texttt{avg\_files\_per\_partition} and the target). File-size heterogeneity
breaks this relationship because the bin-packing filter skips files in
$[96,230]$~MB, forcing the model to learn non-trivial interactions.

This finding has practical implications: any compaction benchmark that uses
only small synthetic files will produce misleadingly perfect ML metrics.
Evaluation must include files spanning the compaction target range to create
non-trivial prediction targets, a concern echoed in LST-Bench's design of
mixed-size workloads~\cite{camachorodriguez2024lstbench}.

\subsubsection{Feature Dominance}
The concentration of importance in files-per-partition features is not a
modelling artefact but reflects the mechanics of bin-packing
compaction~(\ref{eq:rfc}). The compaction procedure~\cite{iceberg_spec} rewrites files smaller
than the minimum threshold (96\,MB) and bins them into target-sized outputs
(128\,MB). The number of eligible small files per partition therefore
determines both the binary outcome (are there files to rewrite?) and the
reduction magnitude.

This dominance has a direct operational consequence:
AutoComp's~\cite{gruenheid2025autocomp} compaction traits already include
partition-level file counts as one of several heuristic inputs. Our
ablation study (Section~\ref{sec:ablation}) provides quantitative evidence
that these traits alone are sufficient for the binary scheduling decision,
while Section~\ref{sec:sensitivity} shows the result holds across model
families (decision trees, random forests). More sophisticated ML models are
only justified when predicting the continuous reduction ratio for
cost-aware scheduling.

\subsubsection{Threats to Validity}
Synthetic UUID payloads compress differently from real columnar
data~\cite{stonebraker2005cstore}.
Partition structure is generated, not domain-derived; query patterns are
not modelled. Spark-local execution may differ from distributed clusters
with HDFS or S3, a gap well-documented in database tuning
research~\cite{vanaken2021inquiry}. The 89.3\% positive class prevalence risks inflating accuracy
metrics. We address this with three mitigations: (i)~a majority-class
baseline (Table~\ref{tab:tpch}), (ii)~prevalence
downsampling experiments showing stable regression at 50\%
balance (Section~\ref{sec:confidence}), and (iii)~bootstrap CIs over
30 random splits (Section~\ref{sec:confidence}). The TPC-H cross-schema experiment
(Section~\ref{sec:tpch}) partially mitigates the synthetic-data concern
by validating on six relational schemas with realistic data types, but
production tables with schema evolution, delete files, and mixed CoW/MoR
strategy~\cite{okolnychyi2024petabyte} remain untested.

Three additional threats deserve explicit acknowledgement.
\emph{Delete files and Merge-on-Read (MoR).}  Our simulation uses
Copy-on-Write exclusively. MoR tables accumulate positional and equality
delete files that compaction must merge into data files ---
a fundamentally different compaction dynamic not captured by our feature
set.
\emph{Sort and z-order compaction.}  We evaluate bin-packing only. Sort-based
compaction rewrites files for query locality; its utility depends on
query workload patterns, which a metadata-only approach cannot capture.
The title ``Predicting Compaction Utility'' therefore applies to
bin-packing utility specifically.
\emph{Temporal evolution.}  Our model operates on static snapshots.
Production tables evolve continuously; a table that does not need
compaction now may need it after the next batch of writes. Temporal
prediction (e.g., forecasting compaction need $k$ commits ahead) is
future work.

\subsubsection{Compaction Target Sensitivity}
All experiments use the same compaction configuration (target 128\,MB,
minimum 96\,MB, maximum 230\,MB). The $k{=}4$ threshold is implicitly
tied to this target: a larger target size (e.g., 512\,MB, Iceberg's
production default) would produce fewer output files per partition,
likely shifting the optimal $k$ downward. The learned XGBoost model
may generalise better than a fixed threshold because it incorporates
multiple features (file sizes, counts, ratios), but this is untested.
Evaluating sensitivity to compaction target size is a priority for
future work.

\subsubsection{Compaction Cost}
Although our framework collects wall-clock compaction duration, this
study focuses on predicting \emph{structural} compaction outcomes
(file count reduction) rather than \emph{economic} cost. A
production cost model would additionally require cloud I/O pricing,
compute cost, and downstream query speedup estimates. We observe a
moderate rank correlation between compaction duration and the number
of files rewritten (Spearman $\rho = 0.76$ on synthetic data,
$\rho = 0.72$ on TPC-H), suggesting that the regression prediction
can serve as a rough proxy for cost ranking, but a full cost-benefit
analysis following Cosine~\cite{chatterjee2022cosine} remains future
work.

\section{Conclusion}

We studied the predictability of bin-packing compaction utility in Apache
Iceberg using 17 metadata features across 2{,}376 synthetic tables and validated
on 96 TPC-H tables spanning six relational schemas. The binary compaction
decision turns out to be a deterministic property of the bin-packing
filter, solvable by a single threshold
(\texttt{max\_files\_per\_partition}~$> 4$) without any learned model.
The genuine ML contribution lies in predicting the continuous
file-reduction ratio: XGBoost achieves $R^2 = 0.998$
(CI~$[0.996, 1.000]$), generalises to TPC-H without retraining
($R^2 = 0.976$), and is robust across prevalence levels, feature
subsets, and model families.

Query benchmarking on 52 TPC-H tables reveals that compaction's downstream
impact is workload-dependent: metadata-heavy queries benefit (1.19$\times$
median \texttt{COUNT(*)} speedup), while full-scan aggregations suffer from
reduced task parallelism (0.45--0.76$\times$). This motivates
workload-aware compaction scheduling: the regression prediction
provides the structural cost signal, but a production system must also
consider query patterns when prioritising compaction.

Future work
includes: (i)~extending the framework to Delta Lake and Hudi;
(ii)~incorporating query-workload features and cloud-cost metrics
following Cosine~\cite{chatterjee2022cosine}; (iii)~multi-table scheduling
via contextual bandits~\cite{marcus2021bao};
(iv)~sort and z-order compaction strategies;
and (v)~validation on production Iceberg catalogs.

\section*{Acknowledgment}
The authors thank the open-source Apache Iceberg and Apache Spark
communities for providing the software infrastructure used in this
research. We thank DSTI School of Engineering for research support.

\bibliographystyle{IEEEtran}
\bibliography{references}

\begin{thebibliography}{10}
\providecommand{\url}[1]{#1}
\csname url@samestyle\endcsname
\providecommand{\newblock}{\relax}
\providecommand{\bibinfo}[2]{#2}
\providecommand{\BIBentrySTDinterwordspacing}{\spaceskip=0pt\relax}
\providecommand{\BIBentryALTinterwordstretchfactor}{4}
\providecommand{\BIBentryALTinterwordspacing}{\spaceskip=\fontdimen2\font plus
\BIBentryALTinterwordstretchfactor\fontdimen3\font minus
  \fontdimen4\font\relax}
\providecommand{\BIBforeignlanguage}[2]{{%
\expandafter\ifx\csname l@#1\endcsname\relax
\typeout{** WARNING: IEEEtran.bst: No hyphenation pattern has been}%
\typeout{** loaded for the language `#1'. Using the pattern for}%
\typeout{** the default language instead.}%
\else
\language=\csname l@#1\endcsname
\fi
#2}}
\providecommand{\BIBdecl}{\relax}
\BIBdecl

\bibitem{armbrust2021lakehouse}
\BIBentryALTinterwordspacing
M.~Armbrust, A.~Ghodsi, R.~Xin, and M.~Zaharia, ``Lakehouse: A new generation
  of open platforms that unify data warehousing and advanced analytics,'' in
  \emph{Proceedings of the 11th Conference on Innovative Data Systems Research
  (CIDR)}, 2021. [Online]. Available:
  \url{http://cidrdb.org/cidr2021/papers/cidr2021_paper17.pdf}
\BIBentrySTDinterwordspacing

\bibitem{iceberg_spec}
{Apache Software Foundation}, ``Apache {Iceberg} table format specification,''
  \url{https://iceberg.apache.org/spec/}, 2024, open table format
  specification, originally developed at Netflix.

\bibitem{armbrust2020delta}
M.~Armbrust, T.~Das, L.~Sun, B.~Yavuz, S.~Zhu, M.~Murthy, J.~Torres, H.~{van
  Hovell}, A.~Ionescu, A.~{\L}uszczak, M.~{{\'S}witakowski},
  M.~Szafra{{\'n}ski}, X.~Li, T.~Ueshin, M.~Mokhtar, P.~Boncz, A.~Ghodsi,
  S.~Paranjpye, P.~Senster, R.~Xin, and M.~Zaharia, ``Delta lake:
  High-performance {ACID} table storage over cloud object stores,''
  \emph{Proceedings of the VLDB Endowment}, vol.~13, no.~12, pp. 3411--3424,
  2020.

\bibitem{hudi_compaction}
{Apache Software Foundation}, ``Apache {Hudi} compaction,''
  \url{https://hudi.apache.org/docs/compaction/}, 2024, documentation of Hudi
  compaction triggers and strategies.

\bibitem{gruenheid2025autocomp}
A.~Gruenheid, J.~Camacho-Rodr{\'i}guez, C.~Curino, R.~Ramakrishnan, S.~Pak,
  S.~Sakdeo, L.~Gandhi, S.~K. Singhal, P.~Nilangekar, and D.~J. Abadi,
  ``{AutoComp}: Automated data compaction for log-structured tables in data
  lakes,'' in \emph{Companion of the 2025 International Conference on
  Management of Data (SIGMOD/PODS '25)}, 2025, pp. 404--417.

\bibitem{databricks_predictive_optimization}
{Databricks}, ``Predictive optimization for {Unity Catalog} managed tables,''
  \url{https://docs.databricks.com/aws/en/optimizations/predictive-optimization},
  2024, aI/ML-driven automatic table maintenance.

\bibitem{snowflake_autoclustering}
{Snowflake Inc.}, ``Automatic clustering,''
  \url{https://docs.snowflake.com/en/user-guide/tables-auto-reclustering},
  2024.

\bibitem{aws_s3tables_compaction}
{Amazon Web Services}, ``How {Amazon S3 Tables} use compaction to improve query
  performance,''
  \url{https://aws.amazon.com/blogs/storage/how-amazon-s3-tables-use-compaction-to-improve-query-performance-by-up-to-3-times/},
  2025.

\bibitem{dayan2017monkey}
N.~Dayan, M.~Athanassoulis, and S.~Idreos, ``Monkey: Optimal navigable
  key-value store,'' in \emph{Proceedings of the 2017 ACM International
  Conference on Management of Data (SIGMOD)}, 2017, pp. 79--94.

\bibitem{dayan2018dostoevsky}
N.~Dayan and S.~Idreos, ``Dostoevsky: Better space-time trade-offs for
  {LSM}-tree based key-value stores via adaptive removal of superfluous
  merging,'' in \emph{Proceedings of the 2018 ACM International Conference on
  Management of Data (SIGMOD)}, 2018, pp. 505--520.

\bibitem{sarkar2021compaction}
S.~Sarkar, D.~Staratzis, Z.~Zhu, and M.~Athanassoulis, ``Constructing and
  analyzing the {LSM} compaction design space,'' \emph{Proceedings of the VLDB
  Endowment}, vol.~14, no.~11, pp. 2216--2229, 2021.

\bibitem{mo2023ruskey}
D.~Mo, F.~Chen, S.~Luo, and C.~Shan, ``Learning to optimize {LSM}-trees:
  Towards a reinforcement learning based key-value store for dynamic
  workloads,'' \emph{Proceedings of the ACM on Management of Data}, vol.~1,
  no.~3, 2023.

\bibitem{ecotune2025}
C.~Luo, S.~Chatterjee, R.~Ketsetsidis, K.~Nagaraj, M.~J. Amiri, and S.~Idreos,
  ``Rethinking the compaction policies in {LSM}-trees,'' \emph{Proceedings of
  the ACM on Management of Data}, 2025, sIGMOD 2025.

\bibitem{chaudhuri2007selftuning}
S.~Chaudhuri and V.~Narasayya, ``Self-tuning database systems: A decade of
  progress,'' in \emph{Proceedings of the 33rd International Conference on Very
  Large Data Bases (VLDB)}, 2007, pp. 3--14.

\bibitem{vanaken2017ottertune}
D.~{Van Aken}, A.~Pavlo, G.~J. Gordon, and B.~Zhang, ``Automatic database
  management system tuning through large-scale machine learning,'' in
  \emph{Proceedings of the 2017 ACM International Conference on Management of
  Data (SIGMOD)}, 2017, pp. 1009--1024.

\bibitem{zhang2019cdbtune}
J.~Zhang, Y.~Liu, K.~Zhou, G.~Li, Z.~Xiao, B.~Cheng, J.~Xing, Y.~Wang,
  T.~Cheng, L.~Liu, M.~Ran, and Z.~Li, ``An end-to-end automatic cloud database
  tuning system using deep reinforcement learning,'' in \emph{Proceedings of
  the 2019 ACM International Conference on Management of Data (SIGMOD)}, 2019,
  pp. 415--432.

\bibitem{pavlo2021noisepage}
A.~Pavlo, M.~Butrovich, L.~Ma, W.~S. Lim, P.~Menon, D.~{Van Aken}, and
  W.~Zhang, ``Make your database system dream of electric sheep: Towards
  self-driving operation,'' \emph{Proceedings of the VLDB Endowment}, vol.~14,
  no.~12, pp. 3211--3221, 2021.

\bibitem{zhou2022dbmeetsai}
X.~Zhou, C.~Chai, G.~Li, and J.~Sun, ``Database meets artificial intelligence:
  A survey,'' \emph{IEEE Transactions on Knowledge and Data Engineering},
  vol.~34, no.~3, pp. 1096--1116, 2022.

\bibitem{marcus2021bao}
R.~Marcus, P.~Negi, H.~Mao, N.~Tatbul, M.~Alizadeh, and T.~Kraska, ``Bao:
  Making learned query optimization practical,'' in \emph{Proceedings of the
  2021 ACM International Conference on Management of Data (SIGMOD)}, 2021, pp.
  1275--1288, best Paper Award.

\bibitem{yang2020qdtree}
Z.~Yang, B.~Chandramouli, C.~Wang, J.~Gehrke, Y.~Li, U.~F. Minhas, P.-{\AA}.
  Larson, D.~Kossmann, and R.~Acharya, ``Qd-tree: Learning data layouts for big
  data analytics,'' in \emph{Proceedings of the 2020 ACM SIGMOD International
  Conference on Management of Data}, 2020, pp. 193--208.

\bibitem{chen2016xgboost}
T.~Chen and C.~Guestrin, ``{XGBoost}: A scalable tree boosting system,'' in
  \emph{Proceedings of the 22nd ACM SIGKDD International Conference on
  Knowledge Discovery and Data Mining}, 2016, pp. 785--794.

\bibitem{tpch2024}
\BIBentryALTinterwordspacing
{Transaction Processing Performance Council}, ``{TPC-H} benchmark
  specification, revision 3.0.1,'' TPC, Tech. Rep., 2024. [Online]. Available:
  \url{https://www.tpc.org/tpch/}
\BIBentrySTDinterwordspacing

\bibitem{jain2023analyzing}
\BIBentryALTinterwordspacing
P.~Jain, P.~Kraft, C.~Power, T.~Das, I.~Stoica, and M.~Zaharia, ``Analyzing and
  comparing lakehouse storage systems,'' in \emph{Proceedings of the 13th
  Conference on Innovative Data Systems Research (CIDR)}, 2023. [Online].
  Available: \url{https://www.cidrdb.org/cidr2023/papers/p92-jain.pdf}
\BIBentrySTDinterwordspacing

\bibitem{camachorodriguez2024lstbench}
J.~Camacho-Rodr{\'i}guez \emph{et~al.}, ``{LST-Bench}: Benchmarking
  log-structured tables in the cloud,'' \emph{Proceedings of the ACM on
  Management of Data}, vol.~2, no.~1, 2024.

\bibitem{stonebraker2005cstore}
M.~Stonebraker, D.~J. Abadi, A.~Batkin, X.~Chen, M.~Cherniack, M.~Ferreira,
  E.~Lau, A.~Lin, S.~Madden, E.~O'Neil, P.~O'Neil, A.~Rasin, N.~Tran, and
  S.~Zdonik, ``{C-Store}: A column-oriented {DBMS},'' in \emph{Proceedings of
  the 31st International Conference on Very Large Data Bases (VLDB)}, 2005, pp.
  553--564.

\bibitem{vanaken2021inquiry}
D.~{Van Aken}, D.~Yang, S.~Brillard, A.~Fiorino, B.~Zhang, C.~Billian, and
  A.~Pavlo, ``An inquiry into machine learning-based automatic configuration
  tuning services on real-world database management systems,''
  \emph{Proceedings of the VLDB Endowment}, vol.~14, no.~7, pp. 1241--1253,
  2021.

\bibitem{okolnychyi2024petabyte}
A.~Okolnychyi, C.~Sun, A.~Tanimura, R.~Spitzer, R.~Blue, S.~Ho, P.~Gu,
  A.~Lakkundi, and G.~Tsai, ``Petabyte-scale row-level operations in data
  lakehouses,'' \emph{Proceedings of the VLDB Endowment}, vol.~17, no.~12, pp.
  4159--4172, 2024.

\bibitem{chatterjee2022cosine}
S.~Chatterjee, M.~Jagadeesan, W.~Qin, and S.~Idreos, ``Cosine: A cloud-cost
  optimized self-designing key-value storage engine,'' \emph{Proceedings of the
  VLDB Endowment}, vol.~15, no.~1, pp. 112--126, 2022.

\end{thebibliography}

\end{document}